\pdfoutput=1

\documentclass[11pt]{article}

\usepackage[preprint]{acl}

\usepackage{times}
\usepackage{latexsym}
\usepackage{multirow}
\usepackage{booktabs}

\usepackage[T1]{fontenc}

\usepackage[utf8]{inputenc}

\usepackage{microtype}

\usepackage{inconsolata}

\usepackage{graphicx}

\title{ReTok: Replacing Tokenizer to Enhance Representation Efficiency \\ in Large Language Model}

\author{Shuhao Gu, Mengdi Zhao, Bowen Zhang, Liangdong Wang, \\ \bf{Jijie Li, Guang Liu}\thanks{* corresponding author} \\
        Data Research Team, BAAI \\ \texttt{\{shgu, liuguang\}@baai.ac.cn}}

\begin{document}
\maketitle
\begin{abstract}



Tokenizer is an essential component for large language models (LLMs), and a tokenizer with a high compression rate can improve the model's representation and processing efficiency. 
However, the tokenizer cannot ensure high compression rate in all scenarios, and an increase in the average input and output lengths will increases the training and inference costs of the model.
Therefore, it is crucial to find ways to improve the model's efficiency with minimal cost while maintaining the model's performance.
In this work, we propose a method to improve model representation and processing efficiency by replacing the tokenizers of LLMs.
We propose replacing and reinitializing the parameters of the model's input and output layers with the parameters of the original model, and training these parameters while keeping other parameters fixed. 
We conducted experiments on different LLMs,
and the results show that our method can maintain the performance of the model after replacing the tokenizer, while significantly improving the decoding speed for long texts.
\end{abstract}

\section{Introduction}
Tokenizer is a basic component of large language models (LLMs)~\cite{DBLP:conf/nips/BrownMRSKDNSSAA20,DBLP:journals/corr/abs-2303-08774,DBLP:journals/corr/abs-2302-13971}, which is used to preprocess text by converting it into a sequence of tokens. 
This process allows the model to handle and analyze text data more efficiently by breaking it down into manageable pieces.
Currently, different LLMs typically use different tokenizers, which are generally trained on their own training corpora with various methods such as BPE~\cite{DBLP:conf/acl/SennrichHB16a}, WordPiece~\cite{DBLP:conf/emnlp/SongSSDZ21}, Unigram~\cite{DBLP:conf/acl/Kudo18}, etc.
Generally, these methods determine the vocabulary of the tokenizer based on the frequency of different tokens in the training data.
This allows the model to obtain an efficient representation of the training data, achieving a relatively high compression rate, i.e., shorter sequence lengths.
However, this also results in a lower compression rate for the tokenizer when there is a significant distribution difference between the test data and the training data~\cite{DBLP:journals/corr/abs-2002-07306,DBLP:journals/corr/abs-2311-08849,DBLP:conf/emnlp/DoblerM23}.
For example, a tokenizer trained on general domain corpora will have a lower compression rate for specific tasks like code and math.
On the one hand, using a low-compression-rate tokenizer will cause the model to output longer sequences during inference, which increases the computational cost and inference time~\cite{DBLP:conf/emnlp/Ahia0GKMST23}.
On the other hand, it will also cause the model to consume more computation during supervised fine-tuning (SFT) on specific languages or tasks, affecting training efficiency and potentially impacting the final performance~\cite{DBLP:journals/corr/abs-2402-01035}.
Besides, if we need to utilize different LLMs in practical applications, e.g., using different LLMs for collaboration~\cite{xiong-etal-2023-examining}, we must optimize and adapt to different tokenizers, which increases development and maintenance costs.
Therefore, substituting pretrained model's original tokenizer with a arbitrary one that provides higher compression rates could substantially reduce the cost of using LLM.



In this work, we propose to \textbf{Re}place \textbf{Tok}nizer (\textbf{ReTok}) of LLM to imporve the model's represenation and processing efficiency, while maintaining the model's performance as much as possible.
First, using the Llama3 tokenizer~\cite{llama3} as an example, which demonstrates high compression rates for English and code, we quickly improve its compression rate for Chinese by expanding the vocabulary.
Then, we propose replacing the model's input and output layers based on the new tokenizer.
Meanwhile, we initialize the new parameters according to the correspondence between the tokens in the new tokenizer's vocabulary and the original tokenizer's vocabulary.
Lastly, these new parameters of the model will be updated, and other parameters will be fixed during model training.
Experiments were conducted on the Qwen1.5-0.5B~\cite{DBLP:journals/corr/abs-2309-16609}, Aquila2-7B~\cite{Aquila2}, and Llama3-8B~\cite{llama3} models. 
The results indicate that our method can maintain the performance of the original LLM after replacing the tokenizer, while significantly reducing the inference time of the model in specific domains.


\section{Related Work}
\begin{figure}
    \centering
    \includegraphics[width=\columnwidth]{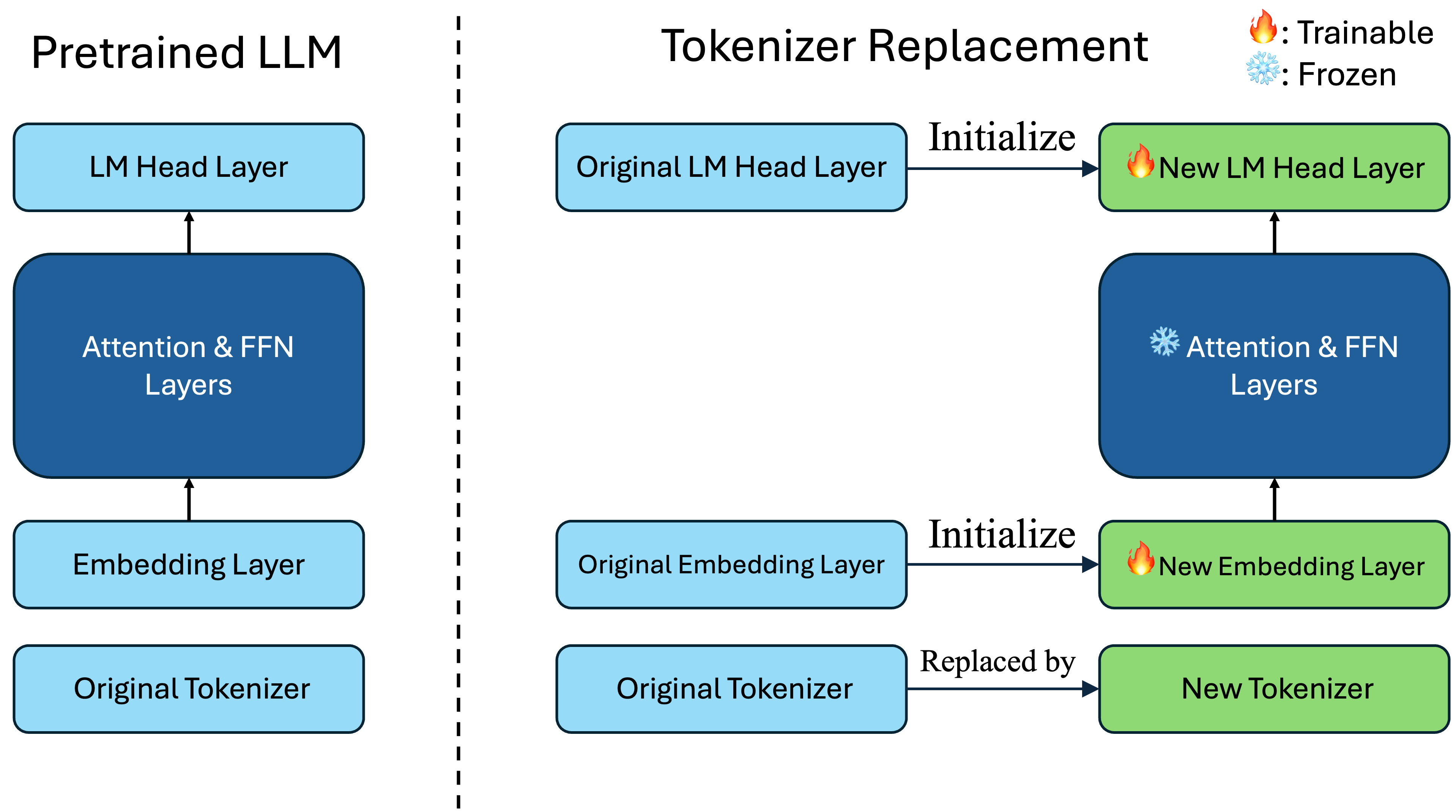}
    \caption{Illustration of the proposed method.}
    \label{fig:method}
\end{figure}

The work on tokenizers can generally be divided into two categories based on the objectives.
The first category of work studies how to design the optimal tokenizer to help the model learn more effectively. 
\citet{DBLP:journals/corr/abs-2402-14903} explored the impact of different tokenization methods for numbers on mathematical problems.
\citet{DBLP:journals/tacl/ClarkGTW22} explored having the model discard the tokenizer to process text, using characters to directly represent the input sequence.
\citet{DBLP:conf/icml/ZhengSJWXXW23} proposed the Syntax-Aware Tokenization method, enhancing the model's capability for code tasks. 
\citet{DBLP:journals/corr/abs-2404-08335} presented a theoretical framework to compare and analyze different tokenization algorithms, and provided several suggestions on how to improve tokenization effectiveness.
Besides, there are also studies focused on improving tokenization effectiveness for some specific languages, such as Arabic~\cite{DBLP:journals/npl/AlyafeaiAGA23} and Japanese~\cite{DBLP:conf/emnlp/TolmachevKK18}.

The second category of work focuses on how to modify the existing tokenizers of the pre-trained model.
\citet{DBLP:journals/tacl/XueBCANKRR22} studied how to let the pretrained model process byte sequences with minimal modifications.
\citet{minixhofer2024zeroshot} proposed to train a hypernetwork taking a tokenizer as input and predicting the corresponding embeddings.
\citet{DBLP:journals/corr/abs-2402-09977} and \citet{DBLP:journals/corr/abs-2402-01035} studied how to adapt the tokenizers of pretrained models to improve efficiency and performance for downstream tasks.

In these works, our work is more similar to the second category of work, where we also improve the tokenizers based on pretrained models. 
Meanwhile, our work focuses more on how to improve the representation efficiency of the model with as little cost as possible while maintaining the overall performance of the model.


\section{Methods}
\begin{table}[t!]
\renewcommand{\arraystretch}{1}
\setlength{\tabcolsep}{2pt}
\begin{tabular}{l|c|cccc|c}
\toprule
\multirow{2}{*}{Tokenizer} & \multirow{2}{*}{Size} & \multicolumn{4}{c|}{Compression Ratio}                           & \multirow{2}{*}{Average} \\ \cmidrule{3-6}
                           &                       & Zh            & En            & Code          & Math          &                          \\ \midrule
Aquila2                    & 100k                  & \textbf{4.70} & 4.42          & 3.20          & 3.77          & 4.02                     \\
Qwen1.5                    & 151k                  & 4.27          & 4.51          & 3.62          & 3.35          & 3.94                     \\
Llama3                     & 128k                  & 3.45          & 4.61          & 3.77          & \textbf{3.88} & 3.93                     \\
ReTok              & 143k                  & 4.60          & \textbf{4.61} & \textbf{3.78} & \textbf{3.88} & \textbf{4.22}            \\ \bottomrule
\end{tabular}
\caption{Compression rates for Chinese, English, code, and math with different tokenizers. ReTok is the tokenizer we developed based on the Llama3 tokenizer, which has a higher overall compression rate.}
\label{tab:tok}
\end{table}

\begin{table*}[t]
\setlength{\tabcolsep}{1.5pt}
\resizebox{2.1\columnwidth}{!}{
\begin{tabular}{l|cccccccccc}
\hline
Model &
  \begin{tabular}[c]{@{}c@{}}PIQA\\ (5-shot)\end{tabular} &
  \begin{tabular}[c]{@{}c@{}}ARC-C\\ (25-shot)\end{tabular} &
  \begin{tabular}[c]{@{}c@{}}HellaSwag\\ (10-shot)\end{tabular} &
  \begin{tabular}[c]{@{}c@{}}MMLU\\ (5-shot)\end{tabular} &
  \begin{tabular}[c]{@{}c@{}}CMMLU\\ (5-shot)\end{tabular} &
  \begin{tabular}[c]{@{}c@{}}AGIEval\\ (0-shot)\end{tabular} &
  \begin{tabular}[c]{@{}c@{}}BBH\\ (3-shot)\end{tabular} &
  \begin{tabular}[c]{@{}c@{}}Human-Eval\\ (0-shot)\end{tabular} &
  \begin{tabular}[c]{@{}c@{}}Gsm8k\\ (5-shot)\end{tabular} &
  Average \\ \hline
Qwen1.5-0.5B & 69.8 & 31.6 & 49.1 & 39.94 & 45.57 & 22.46 & 21.24 & 9.15  & 13.04 & 33.54 \\
+ ReTok      & 70   & 30.6 & 48.4 & 38.74 & 41.32 & 24.38 & 24.83 & 11.59 & 8.64  & 33.17 \\ \hline
Aquila2-7B   & 76.3 & 42.4 & 68.7 & 45.66 & 49.94 & 26.17 & 29.73 & 6.1   & 8.04  & 39.23 \\
+ ReTok      & 76.3 & 42.8 & 68.4 & 43.72 & 41.45 & 23.84 & 22.73 & 14.02 & 6.44  & 37.74 \\ \hline
LLama3-8b    & 82.1 & 59.6 & 82.1 & 66.66 & 50.66 & 26.37 & 61.25 & 26.22 & 50.04 & 56.11 \\
+ ReTok      & 81.9 & 56.4 & 82.1 & 65.65 & 46.35 & 25.53 & 59.92 & 29.27 & 51.4  & 55.39 \\ \hline
\end{tabular}
}
\caption{Results of the main experiments.}
\label{tab:res}
\end{table*}




The proposed method comprises three distinct steps: obtaining a high-compression tokenizer, replacing and initializing the model's input and output layers, and training the model. An illustration of our method is shown in Figure~\ref{fig:method}.

\subsection{Vocabulary Expansion}
First, we need to obtain a high-compression tokenizer, which can be obtained either by training from scratch or by extending an existing tokenizer.
In this work, we exemplify with the Llama3 tokenizer, distinguished for its commendable compression capabilities in English and code, as a paradigm.
By integrating Chinese lexicons into its vocabulary, we enhance its compression efficacy for Chinese text, thereby obtaining a more efficient representation.
Initially, we employ the same pre-tokenization approach as the original tokenizer to segment the Chinese training corpus into distinct chunks.
Subsequently, on this corpus, we utilize byte-level BPE method to learn a new vocabulary and merge it with the original vocabulary to obtain the new vocabulary as well as the tokenizer.
We compared the compression rates of different tokenizers across Chinese, English, code, and math corpora, where the compression rate is computed as:
\begin{equation}
    \mathrm{r} = \frac{byte(s)}{tokenize(s)},
\end{equation}
where $byte(s)$ represents the sequence length when a sentence is represented with bytes, while $tokenize(s)$ represents the sequence length after tokenization.
The results are shown in Table~\ref{tab:tok}, which indicate that the our newly expanded tokenizer (ReTok) achieves the highest overall compression rate.


\subsection{Layer Replacement \& Initialization }

In this step, we replace and reinitialize the input layer (i.e., embedding layer) and the output layer (i.e., LM head layer) of the original model.
To maximize the utilization of the original model's knowledge and accelerate the convergence of the new parameters, we initialize the newly added parameters based on the model's original layers.
In detail, for each token in the new vocabulary, if it is also present in the original vocabulary, we initialize its corresponding new parameters in both the embedding layer and the LM head layer with the parameters from the original model’s layers.
For tokens that do not exist in the original vocabulary, we process these tokens using the original tokenizer to obtain their corresponding ID lists from the original vocabulary. We then retrieve the associated parameters from the original embedding layer and LM head layer, compute their average, and use this average to initialize the new parameters.

\subsection{Model Training}
Then, we choose to train the reinitialized embedding layer and LM head layer of the model, while keeping the main parameters of the model, i.e. attention layers, FFN layers, and layer norm parameters, fixed to prevent catastrophic forgetting.
Given the training data $\mathcal{T} = \{t_0, \ldots, t_n\}$, we use the standard language modeling objective to training the newly added parameters:
\begin{equation}
    \mathcal{L(\mathcal{T})} =\sum_i \log \mathrm{P} (t_i | t_{i-1}, \ldots, t_0; \Theta),
\end{equation}
where the conditional probability $\mathrm{P}$ is modeled using a neural network with parameters $\Theta$. 
Once the model has converged, we can unfreeze the main parameters of the model and conduct joint training on all the parameters, which will further improve the model's performance.


\section{Experiments}




\subsection{Experimental Setups}

\textbf{Data}
To train the reinitialized parameters effectively, it is essential to construct a comprehensive training dataset. 
Ideally, using the original data of the pretrained model would be the best way to restore the model's performance. 
However, in practice, it is often impossible to access the entire pretraining dataset of the pretrained model. 
Therefore, in this work, we use publicly available datasets as much as possible for our experiments.
Our collected training dataset is divided into three parts: Chinese, English, and Code. 
The English data primarily comes from sources such as RedPajama~\cite{together2023redpajama}, Pile~\cite{gao2020pile}, C4~\cite{DBLP:journals/jmlr/RaffelSRLNMZLL20}, and Falcon~\cite{DBLP:conf/nips/PenedoMHCACPAL23}. 
The Chinese data is mainly sourced from Wudao~\cite{DBLP:journals/aiopen/YuanZDDLCZYT21}  and ChineseWebText~\cite{DBLP:journals/corr/abs-2311-01149}. 
The code data is derived from sources like StarCoder~\cite{DBLP:journals/corr/abs-2305-06161}.

\noindent \textbf{Training Details}
We conducted experiments based on Qwen1.5-0.5B~\cite{DBLP:journals/corr/abs-2309-16609}, Aquila2-7B~\cite{Aquila2}, and Llama3-8B~\cite{llama3} models.
The global batch size for training was set to 384K tokens. 
We used a cosine learning rate scheduler with a maximum learning rate of 1e-4 and a warm-up period of 1000 steps.
The training was conducted on 32 NVIDIA A-100 GPUs.

\noindent \textbf{Evaluation} 
The test tasks can be divided into two categories: comprehension tasks and generative tasks. 
The comprehension tasks include PIQA (5-shot)~\cite{bisk2020piqa}, ARC-Challenge (25-shot)~\cite{clark2018think}, HellaSwag (10-shot)~\cite{zellers2019hellaswag}, MMLU (5-shot)~\cite{hendrycks2020measuring}, and CMMLU (5-shot)~\cite{li2023cmmlu}, while the generative tasks include AGIEval (0-shot)~\cite{DBLP:journals/corr/abs-2304-06364}, Big-Bench Hard (BBH) (3-shot)~\cite{DBLP:conf/acl/SuzgunSSGTCCLCZ23}, Human-Eval (0-shot)~\cite{chen2021evaluating}, and Gsm8k (5-shot)~\cite{DBLP:journals/corr/abs-2110-14168}. 
These tasks cover the capabilities in Chinese, English, code, and math.
All tasks were tested using frameworks lm-evaluation-harness~\cite{gao2021harness} and OpenCompass~\cite{2023opencompass}.
All the results were obtained from a single run.

\begin{figure}[t]
    \centering
    \includegraphics[width=\columnwidth]{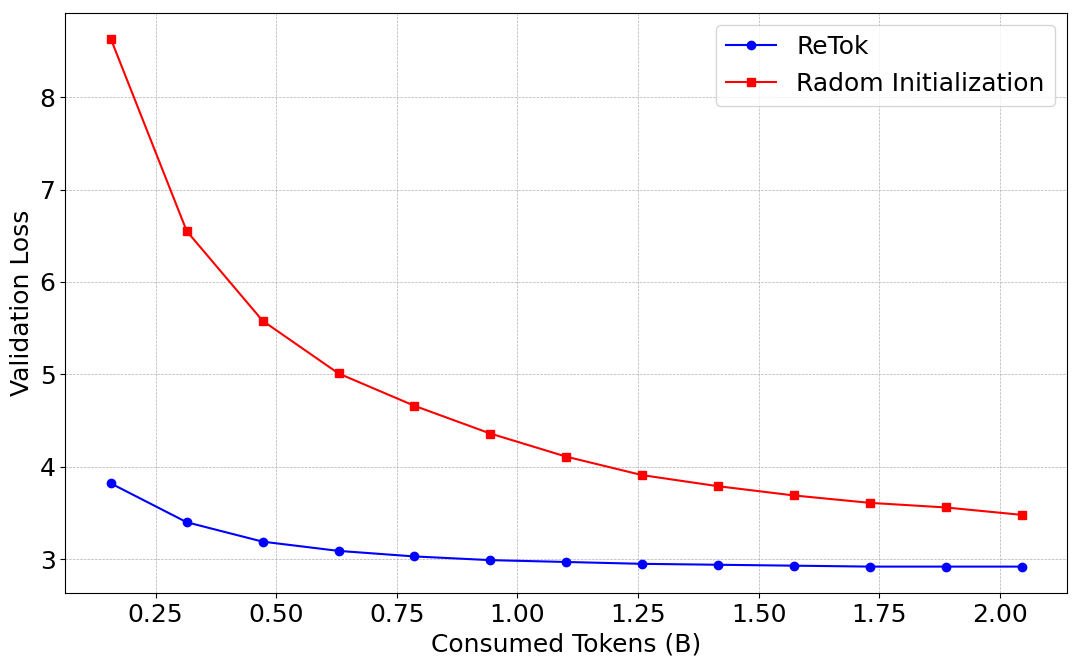}
    \caption{Comparison of validation loss changes between ReTok and random initialization based on Llama3-8B.}
    \label{fig:loss}
\end{figure}

\subsection{Main Results}

The overall results are shown in Table~\ref{tab:res}. 
On most test sets, our method ReTok achieved performance similar to the original models, demonstrating the effectiveness of our approach. 
However, there were notable changes in performance for certain test sets, such as HumanEval and CMMLU. 
This may because that our training data included a significant amount of code content and relatively less Chinese text. 
This phenomenon suggests that the distribution of pre-training data can influence model performance, and our future work will focus on the impact of different data on model performance.

\subsection{Convergence Speed Analysis}

In this section, we analyzed the convergence speed of our method.
We randomly selected a portion of data from the training set and, after a rigorous deduplication, used it as our validation set.
We tested the change of validation loss based on the Llama3-8B model with our method during training. 
Besides, we compared the validation loss changes when the input and output layers of the model were randomly initialized. 
The results are presented in Figure~\ref{fig:loss}. 
The results show that, compared to the commonly used random initialization methods, our method quickly achieves convergence and delivers better performance.

\subsection{Inference Speed Comparison}



\begin{figure}[t]
    \centering
    \includegraphics[width=\columnwidth]{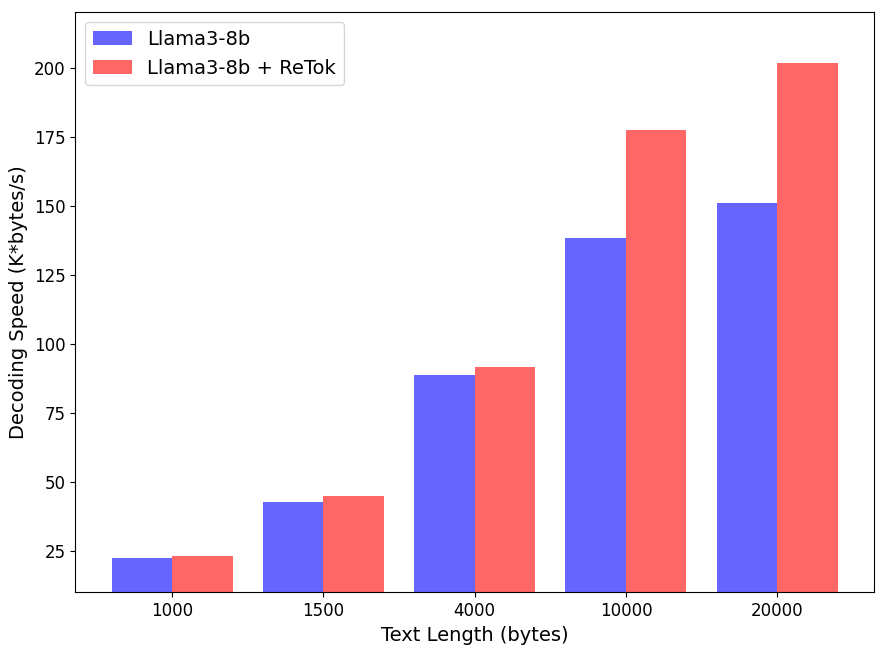}
    \caption{Comparison of decoding speeds based on the Llama3-8B model on the Chinese validation set for texts of different lengths.}
    \label{fig:speed}
\end{figure}
In this experiment, we compared the change in decoding speed of the model before and after tokenizer replacement based on the Llama3-8B model. 
We measured the decoding speed of the model on a Chinese validation set, because our proposed method ReTok primarily improves the compression rate of Chinese.
We grouped the Chinese data based on byte-based sequence lengths and measured the model's speed at each length group.
Then, we used force decoding to process these data, ensuring that the lengths of input and output texts for different models were consistent (but the sequence lengths after tokenization were not the same). 
We compared the decoding speeds of different models, and the results are shown in Figure~\ref{fig:speed}. The results indicate that our method can  improve decoding speed, especially for longer texts.




\section{Conclusion}

In this work, we explored how to improve the model's representation and processing efficiency by replacing the tokenizer of a LLM.
We proposed using the original parameters to initialize the model's input and output layers to accelerate convergence. 
We conducted experiments on multiple test sets, and the results demonstrated that our method can maintain the original performance of the model while increasing decoding speed significantly.

\section*{Limitations}
This work has two main limitations. First, the data we used comes from multiple sources, and we did not analyze the impact of each data source and different data ratios on different models. Second, our test set mainly focuses on the model's capabilities in Chinese, English, code, and math, without testing the model's other capabilities, such as multilingual abilities. We will leave these in future work.

\bibliography{custom}

\end{document}